\documentclass{svproc} 
\usepackage{url}

\usepackage{tabularx}
\usepackage{amsmath}
\usepackage{amssymb}
\usepackage{multirow}
\usepackage{float}
\usepackage[table]{xcolor}
\usepackage{array, makecell}
\usepackage{mathtools}
\definecolor{Gray}{gray}{0.85}
\newcolumntype{g}{>{\columncolor{Gray}}c}
\definecolor{BBL}{rgb}{0.0, 0.0, 0.0}

\makeatletter
\renewcommand*{\@fnsymbol}[1]{\ensuremath{\ifcase#1\or \dagger\or \ddagger\or \mathsection\or \mathparagraph\or \|\or **\or \dagger\dagger\or \ddagger\ddagger\else\@ctrerr\fi}}
\makeatother
\begin{document}
\mainmatter              
\title{Optimizing Feature Selection for Binary Classification with Noisy Labels: A Genetic Algorithm Approach}
%
%
\author{Vandad Imani*\inst{1} \and Elaheh Moradi\inst{1}
Carlos Sevilla-Salcedo\inst{2}\and Vittorio Fortino\inst{3}\and Jussi Tohka\inst{1}\thanks{For the Alzheimer’s Disease Neuroimaging Initiative}}
%
%
\tocauthor{Vandad Imani, Elaheh Moradi, Carlos Sevilla-Salcedo, Vittorio Fortino, and Jussi Tohka}
\institute{A. I. Virtanen Institute for Molecular Sciences, University of Eastern Finland, Finland,\\
\and
Department of Computer Science, Aalto University, Espoo, Finland,\\
\and
Institute of Biomedicine, University of Eastern Finland, Finland,\\
\email{\{vandad.imani, elaheh.moradi, vittorio.fortino, jussi.tohka\}@uef.fi},  
\email{carlos.sevillasalcedo@aalto.fi}
}

\maketitle 

\begin{abstract}
Feature selection in noisy label scenarios remains an understudied topic. We propose a novel genetic algorithm-based approach, the Noise-Aware Multi-Objective Feature Selection Genetic Algorithm (NMFS-GA), for selecting optimal feature subsets in binary classification with noisy labels. NMFS-GA offers a unified framework for selecting feature subsets that are both accurate and interpretable. We evaluate NMFS-GA on synthetic datasets with label noise, a Breast Cancer dataset enriched with noisy features, and a real-world ADNI dataset for dementia conversion prediction.  Our results indicate that NMFS-GA can effectively select feature subsets that improve the accuracy and interpretability of binary classifiers in scenarios with noisy labels. 

\end{abstract}

\section{Introduction}
In the era of big data, machine learning (ML) algorithms are increasingly facing the challenges of high data dimensionality and redundant features. This can make it difficult to extract valuable knowledge from the massive feature space, and can lead to overfitting and poor generalization performance, known as the "curse of dimensionality"~\cite{siblini2019review}. This makes it hard for conventional ML  methods to find valuable information from an expansive feature space laden with irrelevant and redundant features present in data~\cite{pappu2014high}. Feature selection (FS) algorithms can be used to address these challenges by identifying a subset of features that are most relevant to the classification task~\cite{liu2005toward}. These algorithms consider the relationship between features and labels, and depending on this relationship, they can help improve the accuracy and interpretability of ML models.

There are three main types of FS algorithms: filter-based, wrapper-based, and embedded. The filter-based algorithms assess features using an evaluation index that is independent of the learning algorithm~\cite{pan2022multi,santos2016computer,liu1996probabilistic}. Embedded algorithms perform FS as part of the classification algorithm by simultaneously learning the classifier, either within the algorithm or as an added functionality~\cite{sevilla2022multi,imani2021comparison,tuv2009feature}. Wrapper-based algorithms search for the optimal subset of features by evaluating their performance on the classification algorithm~\cite{imani2023multi,mafarja2017hybrid}.

Traditionally, FS algorithms have been tailored for clean datasets, where all the labels in the training set are correct. However, in many practical situations, obtaining a training set with correct labels is difficult, but the labels must be assumed to contain noise. This can be caused by a variety of factors, such as human error, data collection imperfections, or ambiguities in labeling criteria. The presence of noisy labels can severely impact the accuracy and generalization of ML models.

There are a number of methods for addressing noisy labels. One approach is to identify and eliminate samples with noisy labels from the training dataset~\cite{zhu2003eliminating}. Another approach is to design loss functions that are less sensitive to label noise~\cite{wang2019symmetric}. These loss functions can be categorized into two main types: symmetric and asymmetric. Symmetric loss functions~\cite{wang2019symmetric} penalize errors equally, regardless of whether the error is an underprediction or an overprediction, making them less sensitive to label noise when compared to their asymmetric counterparts. In contrast, asymmetric loss functions~\cite{zhou2023asymmetric} may penalize false positives and false negatives differently, potentially leading to higher sensitivity to label noise. A third approach is to correct the loss function based on the estimated label flip rates~\cite{gong2022class,natarajan2013learning}. The label flip rate is the probability that a label is incorrect. By estimating the label flip rate, we can adjust the loss function to be more robust to noise. The work in~\cite{natarajan2013learning} proposed a simple loss function that combines a symmetric loss function and an asymmetric loss function, and the weights of the two loss functions are determined by the estimated label flip rate.

However, FS with noisy labels remains an understudied topic. Addressing this gap, our primary aim in this paper is to introduce and validate a novel approach to FS in noisy label scenarios. Specifically,  
we propose a novel FS approach, called Noise-Aware Multi-objective Feature Selection with Genetic Algorithm (NMFS-GA), based on our recently presented MMFS-GA framework~\cite{imani2023multi} for FS with noisy labels. The algorithm can effectively select informative feature subsets that enhance the accuracy and interpretability of binary classifiers in the presence of noisy labels. The proposed method is evaluated on different datasets to assess its performance. We conducted experiments on synthetic datasets, and applied NMFS-GA to a real-world Breast Cancer dataset that has been enriched with noisy features to simulate challenging conditions. In both synthetic and Breast Cancer datasets, we explored varying degrees of label noise, including 5\%, 10\%, 15\%, and 20\% noise levels, capturing different levels of noise severity. Furthermore, NMFS-GA is evaluated on another real-world ADNI dataset, where we focused on the classification task of stable MCI (sMCI) versus progressive MCI (pMCI). This dataset presents a scenario of asymmetric label noise, which adds to the complexity of the classification problem.

\section{Proposed Method}
In this section, we present the proposed Genetic Algorithm (GA) designed for FS in the context of binary classification with noisy labels (NMFS-GA). We define $\mathbf{D}$ as the underlying joint distribution of a pair of random variables $\{(\mathbf{x}, {y})|\mathbf{x}\in \mathbf{X}$ and $\mathbf{y}\in \mathbf{Y}\} = \mathbf{X} \times \mathbf{Y}$. Here, $\mathbf{X} \subset \mathbb{R}^{d}$ represents the feature space from which the samples are drawn, and $\mathbf{Y} = \{0,1\}$ denotes the output label space. In the ideal noise-free classification scenario, we consider that a training set $S = \{(\mathbf{x}_1,{y}_1), \ldots, (\mathbf{x}_N,{y}_N)\}$ of $N$ samples are drawn independently and identically from $\mathbf{D}$. In this case, all labels $\{{y_i}\}_{i=1}^N$ are correct. However, in real-world classification tasks, noisy labels often exist, leading us to work with a noisy training set $\tilde{S} = {(\mathbf{x}_i,\tilde{{y}}_i)}_{i=1}^N$ obtained from a noisy distribution $\tilde{\mathbf{D}}$.   
Here $\tilde{y}$ is a contaminated version of $y$: 
\begin{equation}
\tilde{y}_i =
\begin{cases}
1 - y_i & \text{with probability $\rho_{noise}$} \\[1pt]
y_i & \text{with probability $(1 - \rho_{noise})$}
\end{cases}.
\end{equation}

Our objective is to use a GA to identify important features in the input data and address the challenges posed by noisy labels to develop effective strategies to learn from the noisy training set, $\tilde{S}$. Specifically, our objective is to identify relevant features in the input data while minimizing the estimated generalization error of the model. To achieve this, we introduce a set of membership indicators, $\mathbf{c} \in \{0,1\}^{1 \times d}$, where $c_j = 1$ indicates the presence of feature $x_j$ in the optimum set of features, while $c_j = 0$ indicates its absence. We define the loss function $\ell : \mathbb{R} \times \mathbf{Y} \rightarrow \mathbb{R}$ to penalize the difference between the model output $h(\mathbf{x})$ and the noisy ground truth label $\tilde{\mathbf{y}}$. The empirical risk of $h$ in the noisy training set $\tilde{\mathbf{S}}$ is represented as:
\begin{equation}
\hat{\mathbf{R}}(h, \tilde{\mathbf{S}}) = \frac{1}{N} \sum_{i=1}^{N} \ell (h(\mathbf{x}_i),\tilde{{y}}_i)
\end{equation}
Ideally, we hope to find an unbiased estimator $\hat{\mathbf{R}}(h, \tilde{\mathbf{S}})$ for $\hat{\mathbf{R}}(h, \mathbf{S})$ given $\tilde{\mathbf{S}}$ so that the adverse impact caused by noisy labels can be removed. In this work, we consider linear $h$ implemened through linear discriminant analysis (LDA). 

\subsection{NMFS-GA Framework}

In this section, we describe a GA-based framework, called NMFS-GA (Noise-Aware Multi-objective Feature Selection Genetic Algorithm), designed to address the FS problem using multiniche techniques. We formulate FS as a multiobjective optimization problem~\cite{deb2002fast}, with the following objective functions:
\begin{equation}
\label{Eq_3:fun}
f_{1}(\mathbf{c}) =  \mathcal{E}(\tilde{y}, h(\mathbf{x};\mathbf{c}));
f_{2}(\mathbf{c}) =  \sum_{j=1}^{d} c_{j}  \\     
\end{equation}
Here, $f_{1}(\mathbf{c})$ measures the classification performance though a specific loss function $\ell$ when only specific features ($c_j = 1$) are considered useful and $f_{2}(\mathbf{c})$ calculates the total number of selected features.  The objective is to simultaneously minimize the classification error while selecting the most informative features. To avoid premature convergence and enhance the identification of important features, NMFS-GA employs multiple niches, each independently evolving its populations through GA operators.

The proposed algorithm consists of the following steps, repeating steps from 2 to 5 until convergence:
\begin{enumerate}
  \item[1.]\textbf{Initialization}: The algorithm randomly generates a population of $C$ chromosomes, each of which represents a potential solution in the form of a feature subset.
  \item[2.]\textbf{Fitness evaluation}: The fitness of each chromosome is evaluated using objective functions in Eq.(\ref{Eq_3:fun}) calculated through 10-fold cross-validation using LDA as the classifier. 
  \item[3.]\textbf{Crossover and Mutation}: The variation operator produces new offspring through crossover or mutation to develop better solutions that will emerge in the population during evolution. In the case of the crossover process, the elements of two solutions of the parental population mate to produce a single offspring. The algorithm uses binomial crossover to generate the offspring since this method is less dependent on the size of the population. In the event of mutation, an element of one solution from the parental population is selected at random and mutated according to the probability rate of mutation to produce a single offspring.
  \item[4.]\textbf{Selection}: The tournament selection operator, adopted by the Nondominated Sorting Genetic Algorithm II~\cite{deb2002fast} as the selection operator, works by randomly selecting two solutions from the population, comparing the solutions with respect to their front ranks and their crowding distance, and selecting the best one.
  \item[5.]\textbf{Migration}: Niches independently evolve their populations through crossover and mutation; nevertheless, niches interact with each other every $5\%$ of the total generations through a genetic operator termed migration, which swaps the top 25\% of their populations.
  \item[6.]\textbf{Termination}: 
  The algorithm terminates after 1000 generations.
\end{enumerate}

\subsection {Loss Functions: Theoretical Analysis}

We can approximate $\mathcal{E}(\tilde{y}, h(\mathbf{x};\mathbf{c})) \approx  \frac{1}{N} \sum_{i=1}^{N} \ell (h(\mathbf{x}_i;\mathbf{c}),\tilde{{y}}_i)$ through various loss functions, summarized in Table~\ref{Tab_1:loss}, which have been proposed to deal with noisy labels. Note that, when needed, LDA outputs classification probabilities, $h(\mathbf{x};\mathbf{c})$ presenting the probability of the class 1.
\textbf{Cross-entropy} loss ($\ell_{CE}$)~\cite{shannon1948mathematical}, known as log loss, is a commonly used loss function for classification tasks, including those with noisy labels~\cite{ghosh2017robust}. It measures the dissimilarity between the predicted probability distribution of the model and the true distribution of the target variables.
\textbf {Symmetric cross-entropy} loss ($\ell_{SCE}$)~\cite{wang2019symmetric}, is a modified version of CE loss function that addresses the issue of imbalanced class distributions. It assigns different weights to positive and negative samples, allowing for more balanced training. \textbf {Generalized cross-entropy} loss ($\ell_{GCE}$)~\cite{zhang2018generalized} loss function is a flexible variant of the CE loss that encompasses multiple loss functions by introducing a parameter, $q$. It generalizes the standard cross-entropy loss and allows for various levels of optimization. \textbf {Joint optimization} loss ($\ell_{JOL}$)~\cite{tanaka2018joint}, is designed to streamline model training by optimizing multiple objectives simultaneously. It consists of three key components: $\ell_{c}$ measures the Kullback-Leibler divergence loss, assessing the difference between predicted probabilities and noisy labels; $\ell_{p}$ acts as a penalty to encourage accurate probability predictions; and $\ell_{e}$ penalizes significant deviations in model parameters to prevent overfitting.
\textbf {Peer loss function} ($\ell_{PL})$~\cite{liu2020peer} is a robust loss function that can be used to learn from noisy labels without knowing the noise rate. The peer loss function penalizes the model for making different predictions for two similar examples. \textbf {Class-Wise denoising loss function} ($\ell_{CWD}$)~\cite{gong2022class} is a robust loss function that can be used to learn from noisy labels in a class-wise manner. The CWD loss function first estimates the centroid of each class in the training set. Then, it penalizes the model for making predictions that are far from the centroid of the class. In addition, we use \textbf{balanced accuracy} ($\ell_{BA}$) as our baseline metric to evaluate loss function performance with noisy labels.
\begin{table}[H]
\caption{Summary of Loss Functions, we define $\tilde{\iota} = 1 - \tilde{y}$, $\psi = 1- h(\mathbf{x};\mathbf{c})$ and $a = 1-\alpha$.}
\scriptsize 
\renewcommand{\arraystretch}{1.5}
\begin{tabularx}{\linewidth}{c c X}
\hline
\hline
\textbf{Loss Function} & \textbf{Formula} & \textbf{Description} \\
\hline

Cross-entropy & $- [\tilde{y} \log(h(\mathbf{x};\mathbf{c})) + \tilde{\iota} \log(\psi)]$& Measures the dissimilarity between the predicted and true probability distributions.\\
\makecell{Symmetric cross-\\ entropy} & $- \left(\alpha \tilde{y} \log(h(\mathbf{x};\mathbf{c})) + a(\tilde{\iota}) \log(\psi)\right)$&$\alpha$ is a balancing parameter between positive and negative samples. \\
\makecell{Generalized\\ cross-entropy} & $  \tilde{y} \frac{1 - h(\mathbf{x};\mathbf{c})^q}{q} + \tilde{\iota} \frac{(1 - \psi^q)}{q} $&$q>0.5$ more noise tolerance, slower convergence; $q<0.5$ Faster convergence, less noise tolerance. \\

Joint optimization & $\ell_{c}(h(\mathbf{x}_i;\mathbf{c}),\tilde{{y}}_i)+\alpha \ell_{p}(h(\mathbf{x};\mathbf{c})) + \beta \ell_{e}(h(\mathbf{x};\mathbf{c}))$& $\alpha$ balances loss and prediction penalty; $\beta$ balances between predicted probabilities and model parameters.\\

Peer loss & $\ell(f(x_n),\tilde{y}_n) - \ell(f(x_{n1}),\tilde{y}_{n2})$& $\ell(.)$ is 0-1 loss; Two randomly sampled peer samples are $(x_n,\tilde{y}_n)$ and $(x_{n1},\tilde{y}_{n2})$. \\

\makecell{Class-Wise\\ denoising} & $ h(\mathbf{x};\mathbf{c})^2 + 1 + Q\langle\tilde{\mu},\mathbf{w} \rangle $& For squared loss, $Q$ is a constant value, and $\tilde{\mu}$ is a dataset centroid estimate, $\mathbf{w}$ are LDA coefficients.\\

\hline
\hline
\end{tabularx}
\label{Tab_1:loss}
\vspace{-2em}
\end{table}

\section{Experimental Results}
We evaluated the proposed NMFS-GA algorithm on four different datasets, two synthetic and two real world data sets.

\subsection{Synthetic Data}
We constructed two synthetic 500-dimensional datasets (A and B) similar to that of~\cite{kvrivzek2007improving} with three simulated label noise rates by flipping the true label with a probability of 0.05, 0.1, or 0.15. The data is structured in a way that it is challenging for feature-ranking or greedy FS methods to find an optimal feature subset.
Specifically, in dataset A, only 6 features exhibit discriminative power, while the other 494 features are non-informative. Moreover, dataset B is designed with 7 discriminative features and 493 non-informative features. The informative features of the samples in each dataset have the same covariance matrix but with different means, and the classes are normally distributed. For dataset A, LDA achieved an accuracy of 72\% for dataset A and 69\% for dataset B without label noise. The Bayes error rates for datasets A and B obtaining values of 0.046 and 0.141, respectively. We evaluate the actual probability of correct classification (PCC) under the assumption of an infinite number of (noiseless) test samples by using conditional PCC as a performance measure. The conditional PCC was computed with the Monte Carlo integration with 10 million simulated test samples.

\paragraph{\textbf{Comparison of loss functions.}}
Table~\ref{Tab_2:Binary_SynData} presents the performance comparison of various loss functions under different label noise rates. The results with the dataset A indicate that all loss functions can achieve PCCs that are close to the optimal (1 - Bayes error rate) when 5\% of the labels are noisy. However, PCCs decrease as the noise ratio increases, which is expected. The CWD and BA losses prove the most robust, consistently achieving the highest PCCs. The other loss functions  are less noise-resistant, with mean PCCs ranging from 0.84 to 0.87 at 5\% noise. At 10\% noise, CWD, BA, and GCE maintain their PCC at 0.79, while others drop to 0.70-0.75. Notably, as the noise ratio escalated to 15\%, the BA achieved a mean a PCC  of 0.70, closely followed by CWD and GCE loss functions.

The comparison of results with the dataset B across different noise levels reveals insights into loss function performance. At 5\% noise, most loss functions perform similarly, the CWD achieved an average PCC of 78\%. At this level, the choice of loss function has minimal impact. At 10\% noise, CWD maintains its competitiveness (PCC 72\%), followed by BA (PCC 71\%), with others showing little difference. As noise rises to 15\%, the impact of label noise on accuracy becomes clear. The CWD remains robust with PCC of 61\%, but all loss functions suffer reduced PCC, indicating the challenge posed by substantial label noise.

Across all noise levels (5\%, 10\%, and 15\%), mean PCC differences between loss functions are not substantial. BA, CWD, and GCE consistently demonstrate similar performance, suggesting that the choice of loss function may have limited impact on classification accuracy when using the NMFS-GA algorithm in scenarios with varying label noise. Given this consistent performance, these three loss functions—BA, CWD, and GCE—were selected for the subsequent experiments. 
\vspace{0mm}
\begin{table}[h]
\footnotesize
\begin{center}
\caption{Comparison of different loss functions used in NMFS-GA algorithm with synthetic data. The table displays the average PCC and its standard deviation across 10 experiments for each loss function under three different label noise rates. There were 100 training samples per class.}
\setlength{\tabcolsep}{5.5pt} 
\renewcommand{\arraystretch}{2.0} 
\begin{tabular}{l c c c c c c c}
\hline
\hline
\multirow{2}{*}{Experiments}  & \multirow{2}{*}{Noise rate} & \multicolumn{6}{c}{Accuracy} \\ \cline{3-8}  
 & $\rho_{noise}$ & BA & CWD & SCE & GCE & JOL & PL\\

 \hline

\multirow{3}{*}{Dataset \: A}  & 0.05 & \makecell{0.87$\pm$\\0.039}  & 
\textbf{ \makecell{0.88$\pm$\\0.039}}  & \makecell{0.85$\pm$\\0.031}  & \makecell{0.87$\pm$\\0.040}  & \makecell{0.84$\pm$\\0.032}  & \makecell{0.86$\pm$\\0.026}  \\

& 0.1  & 
\textbf{\makecell{0.79$\pm$\\0.063}}  & 
\textbf{\makecell{0.79$\pm$\\0.049} } & \makecell{0.73$\pm$\\0.075}  & \textbf{\makecell{0.79$\pm$\\0.085}}  & \makecell{0.75$\pm$\\0.070}  & \makecell{0.70$\pm$\\0.068}  \\
& 0.15  & \textbf{\makecell{0.70$\pm$\\0.060}}  & 
\makecell{0.65 $\pm$\\ 0.064}  & \makecell{0.61 $\pm$\\ 0.038}  & \makecell{0.65 $\pm$\\ 0.096}  & \makecell{0.62 $\pm$\\ 0.089}  & \makecell{0.60 $\pm$\\ 0.063}  \\
\hline

\multirow{3}{*}{Dataset \: B}   & 0.05 & \makecell{0.77$\pm$\\0.028}  & 
\textbf{\makecell{0.78$\pm$\\0.044}}  & \makecell{0.77$\pm$\\0.045}  & \makecell{0.77$\pm$\\0.033}  & \makecell{0.73$\pm$\\0.026}  & \makecell{0.73$\pm$\\0.047}  \\
& 0.1 & \makecell{0.71$\pm$\\0.034}  & 
\textbf{\makecell{0.72$\pm$\\0.048}}  & \makecell{0.68$\pm$\\0.061}  & \makecell{0.70$\pm$\\0.034}  & \makecell{0.63$\pm$\\0.082}  & \makecell{0.68$\pm$\\0.048}  \\
& 0.15 & \makecell{0.58 $\pm$\\ 0.074}  & 
\textbf{\makecell{0.61 $\pm$\\ 0.055}}  & \makecell{0.56 $\pm$\\ 0.040}  & \makecell{0.58 $\pm$\\ 0.052}  & \makecell{0.59 $\pm$\\ 0.032}  & \makecell{0.58 $\pm$\\ 0.050}\\

\hline

\hline

\hline

\end{tabular}
\label{Tab_2:Binary_SynData}
\end{center}
\vspace{0mm}
\end{table}

\paragraph{\textbf{Effect of Sample Size on Algorithm Robustness}}. Table~\ref{Tab_3:Binary_SynData_50_100} explores the robustness of NMFS-GA algorithms with reduced dataset sizes (25 and 50 instances per class).  
As sample size decreases, the PCC of all loss functions tends to decline. Despite smaller datasets of type A, our NMFS-GA algorithm, utilizing GCE loss, maintains competitiveness with an average PCC of 64\%, closely followed by CWD (PCC 63\%) at a 10\% noise ratio. Experiments with type B dataset (Bayes error rate of 0.141) also show CWD achieving 62\% accuracy. At a 15\% noise ratio with a dataset size of 50 samples per class, Table~\ref{Tab_3:Binary_SynData_50_100} reveals a drop in classification accuracy, reflecting the challenges posed by substantial label noise. CWD achieved a mean accuracy of 52\% and 57\%, while GCE reaches 58\% and 57\% for higher and lower discriminative power datasets, respectively.

Increasing the sample size to 50 instances per class yields substantial accuracy improvements across all methods. In Table~\ref{Tab_3:Binary_SynData_50_100}, for experiments with 50 samples per class, GCE loss function achieves an average accuracy of 73\% for higher discriminative power, closely followed by CWD at 72\% at 10\% noise. In the lower discriminative power dataset, CWD loss function reaches 70\% PCC, with GCE at 69\%.

Turning to the 15\% noise ratio results, a consistent decline in PCC is observed for all loss functions due to the increased noise levels. In the dataset with higher discriminative power, the average PCC for GCE loss function is 65\%, while CWD achieved PCC of 59\%. In the dataset with lower discriminative power, CWD loss function achieved average accuracy of 63\%, followed by GCE at 62\%.
\begin{table}[h]
\footnotesize
\begin{center}
\caption{Comparison of loss functions with 25 and 50 samples per class. The table displays the average PCC and its standard deviation across 10 experiments for each loss function.}

\setlength{\tabcolsep}{5.0pt} 
\renewcommand{\arraystretch}{2.0} 
\begin{tabular}{l c g c g c g c}
\hline
\hline
\multirow{2}{*}{Experiment}  & \multirow{2}{*}{Noise rate} & \multicolumn{2}{c}{BA} &  \multicolumn{2}{c}{CWD} & \multicolumn{2}{c}{GCE} \\\cline{3-8}

 & $\rho_{noise}$ & 50 & 100 & 50 & 100 & 50 & 100 \\
 \hline

\multirow{2}{*}{Dataset \: A}   & 0.1 & \makecell{ 0.57 $\pm$\\ 0.080 } & \makecell{ 0.64 $\pm$\\ 0.084 } & \makecell{ 0.63 $\pm$\\ 0.108 }& \makecell{ 0.72 $\pm$\\ 0.022 } & \textbf{\makecell{ 0.64 $\pm$\\ 0.096 }}& \textbf{\makecell{ 0.73 $\pm$\\ 0.052 }}  \\
& 0.15 &\makecell{ 0.50 $\pm$\\ 0.000 } & \makecell{ 0.60 $\pm$\\ 0.070 }&
\makecell{ 0.52 $\pm$\\ 0.070 } &\makecell{ 0.59 $\pm$\\ 0.094 } &\textbf{\makecell{ 0.58 $\pm$\\ 0.089 }} &\textbf{\makecell{ 0.65 $\pm$\\ 0.103 }} \\
\hline                                  
\multirow{2}{*}{Dataset \: B}   & 0.1 & \makecell{ 0.56 $\pm$\\ 0.079 } & \makecell{ 0.64 $\pm$\\ 0.031 } & 
\textbf{\makecell{ 0.62 $\pm$\\ 0.102 }}& \textbf{\makecell{ 0.70 $\pm$\\ 0.032 }}  & \makecell{ 0.58 $\pm$\\ 0.089 } & \makecell{ 0.69 $\pm$\\ 0.039 }  \\
& 0.15 & \makecell{ 0.51 $\pm$\\ 0.043 }& \makecell{ 0.58 $\pm$\\ 0.071 }& 
\textbf{\makecell{ 0.57 $\pm$\\ 0.084 }} &\textbf{\makecell{ 0.63 $\pm$\\ 0.110 }} & \textbf{\makecell{ 0.57 $\pm$\\ 0.095 }}&\makecell{ 0.62 $\pm$\\ 0.083 } \\
\hline

\hline

\hline

\end{tabular}
\label{Tab_3:Binary_SynData_50_100}
\end{center}
\vspace{0mm}
\end{table}

\subsection{Breast Cancer Dataset}

The publicly available breast cancer dataset~\cite{misc_breast_cancer_wisconsin_(diagnostic)_17} comprises 569 samples of which 212 represent malignant and 357 benign cases. Each sample has 30 features, and we further added 300 noise features, drawn from the Gaussian distribution. The features were randomly permuted to prevent any FS algorithm from learning the feature order. To assess the impact of label noise on the performance of our model, we employed 10-fold cross-validation and experiments with two simulated noise rates adding label noise to the training set, flipping the true label with a probability of 0.1  or 0.2.

In table~\ref{Tab_4:BCD}, we compare the classification performance of various loss functions in the NMFS-GA algorithm at noise levels of 10\% and 20\%. At 10\% noise, all three loss functions achieve similar balanced accuracy values, approximately ranging from 0.85 to 0.88. Even at 20\% noise, they perform well, with balanced accuracy values 
 of 0.72 to 0.77. CWD consistently outperforms others, followed by BA, and GCE performs slightly worse but still reasonably well. These findings underline the effectiveness of the NMFS-GA algorithm in handling noisy data using various loss functions.
\begin{table}[h]

\footnotesize
\begin{center}
\caption{Performance comparison of the NMFS-GA Algorithm using different Loss Functions, across two distinct noise rates (10\% and 20\%). The evaluated metrics were balanced accuracy, sensitivity (SEN), specificity (SPE), and area under the curve (AUC), each represented as the mean value across the folds of 10-fold cross-validation $\pm$  standard deviation.}

\setlength{\tabcolsep}{3.2pt} 
\renewcommand{\arraystretch}{2.0} 
\begin{tabular}{l gc gc gc gc}
\hline
\hline
\multirow{2}{*}{Method/}   & \multicolumn{2}{c}{\makecell{Balanced \\ Accuracy
}} & \multicolumn{2}{c}{SEN} & \multicolumn{2}{c}{SPE} & \multicolumn{2}{c}{AUC} \\\cline{2-9} 

$\rho_{noise}$ & 0.1 & 0.2 & 0.1 & 0.2 & 0.1 & 0.2 & 0.1 & 0.2 \\

 \hline

$BA$ & \makecell{0.87 $\pm$\\ 0.061} & \makecell{0.76 $\pm$\\ 0.074} & \makecell{ 0.75 $\pm$\\ 0.113 } & \makecell{ 0.55 $\pm$\\ 0.129 } & \makecell{ 0.99 $\pm$\\ 0.028 } & \makecell{ 0.96 $\pm$\\ 0.043 } & \textbf{\makecell{ 0.98 $\pm$\\ 0.025 }} & \textbf{\makecell{ 0.90 $\pm$\\ 0.063 }}\\

$CWD$ & \textbf{\makecell{0.88 $\pm$\\ 0.044}} & \textbf{\makecell{0.77 $\pm$\\ 0.039}}& \makecell{ 0.75 $\pm$\\ 0.088 } & \makecell{ 0.58 $\pm$\\ 0.087 } & \makecell{ 1.00 $\pm$\\ 0.000 } & \makecell{ 0.97 $\pm$\\ 0.030 } & \textbf{\makecell{ 0.98 $\pm$\\ 0.017 }} & \makecell{ 0.89 $\pm$\\ 0.058 }\\
$GCE$ & \makecell{0.85 $\pm$\\ 0.056} & \makecell{0.72 $\pm$\\ 0.044}& \makecell{ 0.73 $\pm$\\ 0.115 } & \makecell{ 0.53 $\pm$\\ 0.076 } & \makecell{ 0.97 $\pm$\\ 0.022 } & \makecell{ 0.90 $\pm$\\ 0.045 } & \makecell{ 0.95 $\pm$\\ 0.025 } & \makecell{ 0.82 $\pm$\\ 0.046 }\\

\hline

\hline

\hline

\end{tabular}
\label{Tab_4:BCD}
\end{center}
\vspace{0mm}
\end{table}

\subsection{ADNI data}
 
We used  ADNI data (\url{http://adni.loni.usc.edu}) to predict future dementia in participants experiencing Mild Cognitive Impairment (MCI)~\cite{moradi2015machine} as a real-world test-bed for our algorithm. The ADNI was launched in 2003  as a public-private partnership, led by Principal Investigator Michael W. Weiner, MD. 
For up-to-date information, see (\url{www.adni-info.org}).

We used volumes of specific brain structures, extracted from magnetic resonance images (MRI,) as features. 
MRIs were processed through the CAT12 pipeline \cite{gaser2022cat}. Processed MRIs were averaged into regional volumes according to the neuromorphometrics atlas, which is based on MR images from 30 subjects from the OASIS database with 138 manually annotated structures provided by Neuromorphometrics, Inc. \url{(http://neuromorphometrics.com)} \cite{marcus2007open}.
The MCI participants were categorized into two groups according to their follow-up diagnosis: stable MCI (sMCI) and progressive MCI (pMCI). The sMCI group consisted of 191 participants with consistent MCI diagnosis and available follow-up diagnosis at least for 5 years. To introduce label noise we also included 189 participants in sMCI group with available follow-up diagnoses for only 3 years. For the pMCI group, we included all MCI participants who converted to dementia during the available follow-up period. Within this category, we identified a consistent pMCI subgroup consisting of 261 participants who consistently received a dementia diagnosis in their last two available follow-up diagnoses. Additionally, there was a noisy pMCI group consisting of 112 pMCI participants who received a dementia diagnosis only in their last available follow-up diagnosis.

According to table~\ref{Tab_4:ADNI}, all four loss functions performed well on the ADNI dataset, with balanced accuracy values
ranging from 0.76 to 0.80. This suggests that the NMFS-GA algorithm, regardless of the specific loss function used, is effective in achieving a reasonable balance between sensitivity and specificity. 
The sensitivity and specificity metrics, which quantify the ability to correctly identify true positive and true negative cases, respectively, range from 0.73 to 0.77 and 0.78 to 0.84 across different loss functions. This indicates that the algorithm performs consistently well in predicting future dementia in MCI patients by classification of pMCI vs. sMCI groups. The GCE achieved the highest Balanced Accuracy value, followed by the CWD. 

\begin{table}[h]

\footnotesize
\begin{center}
\caption{Performance comparison of the NMFS-GA algorithm using different loss functions for sMCI vs. pMCI on the ADNI dataset. The evaluated metrics include balanced accuracy, sensitivity (SEN), specificity (SPE), and area under the curve (AUC), each represented as the mean value across the folds of 10-fold cross-validation $\pm$  standard deviation.}

\setlength{\tabcolsep}{7.0pt} 
\renewcommand{\arraystretch}{2.0} 
\begin{tabular}{l c c c c}
\hline
\hline
sMCI\: vs\: pMCI & & & &  \\ \cline{1-5}
Method   & \makecell{Balanced \\ Accuracy} & SEN & SPE & AUC \\ \cline{1-5}     

 \hline

 BA & 0.77 $\pm$ 0.045  & 0.73 $\pm$ 0.063  & 0.81 $\pm$ 0.097 & 0.84 $\pm$ 0.047  \\
CWD & 0.78 $\pm$ 0.036 & 0.74 $\pm$ 0.042 & 0.82 $\pm$ 0.045 & 0.85 $\pm$ 0.036\\
GCE & 0.80 $\pm$ 0.052  & 0.77 $\pm$ 0.085 & 0.84 $\pm$ 0.077 &  0.86 $\pm$ 0.050\\

\hline

\hline

\hline

\end{tabular}
\label{Tab_4:ADNI}
\end{center}
\vspace{0mm}
\end{table}

\section{Conclusion}

We investigated the efficacy of the Feature Selection Genetic Algorithm (NMFS-GA) for binary classification tasks with noisy labels, employing various loss functions. Our experiments consisted of synthetic datasets with varying levels of label noise, the Breast Cancer dataset augmented with noise features, and a real-world ADNI dataset for the classification of stable Mild Cognitive Impairment (MCI) versus converter Dementia (CDE). NMFS-GA, equipped with multiple loss functions, demonstrated notable robustness to label noise. It consistently outperformed across different noise ratios, underscoring its reliability in handling noisy data. Across all datasets and noise conditions, NMFS-GA consistently achieved competitive classification accuracy. Variations in the choice of loss function did not lead to drastic differences in performance, indicating the algorithm's versatility and effectiveness. Our experiments revealed that as sample sizes decreased, classification accuracy naturally decreased. However, NMFS-GA remained effective even with limited training data, highlighting its adaptability and potential for various applications.

\subsubsection*{Acknowledgments}
Supported by grants  346934, 332510 from the Academy of Finland; grant 351849 from the Academy of Finland under the frame of ERA PerMed ("Pattern-Cog"); Sigrid Juselius Foundation; grant 220104 from Jenny ja Antti Wihurin; grant 65221647 from Pohjois-Savon Rahasto; and the Doctoral Program in Molecular Medicine (DPMM) from the University of Eastern Finland. The computational analyses were performed on servers provided by UEF Bioinformatics Center, University of Eastern Finland, Finland.

Data collection and sharing regarding ADNI data was funded by the Alzheimer's Disease Neuroimaging Initiative (ADNI) (National Institutes of Health Grant U01 AG024904) and DOD ADNI (Department of Defense award number W81XWH-12-2-0012). ADNI is funded by the National Institute on Aging, the National Institute of Biomedical Imaging and
Bioengineering, and through generous contributions from the following: AbbVie, Alzheimer’s Association; Alzheimer’s Drug Discovery Foundation; Araclon Biotech; BioClinica, Inc.; Biogen; Bristol-Myers Squibb Company; CereSpir, Inc.; Cogstate; Eisai Inc.; Elan Pharmaceuticals, Inc.; Eli Lilly and Company; EuroImmun; F. Hoffmann-La Roche Ltd and its affiliated company Genentech, Inc.; Fujirebio; GE Healthcare; IXICO Ltd.; Janssen Alzheimer Immunotherapy Research \& Development, LLC.; Johnson \& Johnson Pharmaceutical Research \& Development LLC.; Lumosity; Lundbeck; Merck \& Co., Inc.; Meso Scale Diagnostics, LLC.; NeuroRx Research; Neurotrack Technologies; Novartis Pharmaceuticals Corporation; Pfizer Inc.; Piramal Imaging; Servier; Takeda Pharmaceutical Company; and Transition Therapeutics. The Canadian Institutes of Health Research is providing funds to support ADNI clinical sites in Canada. Private sector contributions are facilitated by the Foundation for the National Institutes of Health (www.fnih.org). The grantee organization is the Northern California Institute for Research and Education, and the study is coordinated by the Alzheimer’s Therapeutic Research Institute at the University of Southern California. ADNI data are disseminated by the Laboratory for Neuro Imaging at the University of Southern California.

%


\end{document}